\documentclass[runningheads]{llncs}
\usepackage[T1]{fontenc}
\usepackage{graphicx}
\usepackage{booktabs}
\usepackage[misc]{ifsym}
\newcommand{\corr}{(\Letter)}
\usepackage{amsmath,amsfonts,amssymb,dsfont,bm}
\usepackage{xcolor}
\usepackage{algorithm}
\usepackage{algorithmic}
\usepackage{hyperref}
\usepackage{multirow}
\usepackage{ulem}

\begin{document}

\title{Classification Tree-based Active Learning: A Wrapper Approach}

\titlerunning{Classification Tree-based Active Learning}

\author{Ashna Jose\inst{1} \corr \and
Emilie Devijver\inst{2} \and
Massih-Reza Amini\inst{2} \and
Noel Jakse \inst{1}\and
Roberta Poloni \inst{1}
}
\authorrunning{A. Jose et al.}
\institute{SIMaP, Grenoble-INP, CNRS, University of Grenoble Alpes, 38042 Grenoble, France 
\and
LiG, Grenoble-INP, CNRS, University of Grenoble Alpes, 38042 Grenoble, France
\email{ashna.jose@grenoble-inp.fr}}
\maketitle

\begin{abstract}

Supervised machine learning often requires large training sets to train accurate models, yet obtaining large amounts of labeled data is not always feasible.
Hence, it becomes crucial to explore active learning methods for reducing the size of training sets while maintaining high accuracy. The aim is to select the optimal subset of data  for labeling from an initial unlabeled set, ensuring precise prediction of outcomes. However, conventional active learning approaches are comparable to classical random sampling. This paper proposes a wrapper active learning method for classification, organizing the sampling process into a tree structure, that improves state-of-the-art algorithms. A classification tree constructed on an initial set of labeled samples is considered to decompose the space into low-entropy regions. Input-space based criteria are used thereafter to sub-sample from these regions, the total number of points to be labeled being decomposed into each region. 
This adaptation proves to be a significant enhancement over existing active learning methods. Through experiments conducted on various benchmark data sets, the paper demonstrates the efficacy of the proposed framework by being effective in constructing accurate classification models, even when provided with a severely restricted labeled data set.

\end{abstract}

\section{Introduction}

Creating an optimal training set remains a fundamental challenge in machine learning, despite significant progress in recent years. With the widespread adoption of machine learning across various disciplines, the cost associated with labeling observations has become a significant concern in many practical applications. However, as passive sampling reflects the distribution over the output, it may provide redundant information.  This realization highlights the critical importance of selecting the most informative data points to enhance the efficacy of machine learning models, a topic that lies at the heart of active learning methodologies. Active learning \cite{survey4}, particularly in the pool-based setting \cite{pool_based}, assumes access to a large pool of unlabeled data samples. Through iterative selection of the most informative points, typically guided by the prediction performance or a pre-defined budget, active learning aims to maximize the efficiency of the labeling process. While active learning for classification tasks has seen more development compared to regression, the field still lacks a universally effective method that consistently outperforms standard random sampling.

Active learning methods for classification, designed to smartly select informative data points for labeling, can broadly be categorized into two classes: model-free and model-based approaches. Model-free approaches rely solely on covariate information, while model-based approaches utilize a classifier trained on a subset of labeled data points to guide the selection of subsequent points for labeling.

Among model-free approaches, Greedy Sampling over the feature space (GSx, \cite{gsx}) and Iterative Representativeness Diversity Maximization (iRDM, \cite{irdm}), prioritize the selection of diverse points across the feature space to ensure comprehensive exploration. GSx focuses on geometric distances to select diverse points, while iRDM introduces a trade-off between representativeness and diversity by leveraging centroids generated by k-means clustering. Diversity and representativity-based sampling strategies aim to select instances that well represent the overall pattern of the unlabeled data pool. Clustering strategies, based on hierarchical clustering \cite{hier} and k-Center \cite{kcenter}, leverage cluster information to calculate representativeness and minimize distances to cluster centers, respectively. 

On the other hand, model-based methods utilize the classifier's predictions to identify relevant points for labeling. Uncertainty Sampling (US, \cite{us}) and its variants, such as Least Confident \cite{least_confidence}, Margin-based \cite{margin}, and Entropy-based \cite{entropy}, query instances with the least certainty in their predicted labels. While simple and computationally efficient, US may overlook the distribution of classes, potentially impacting sampling quality. Expected Model Change (EMC, \cite{emcm}) selects instances that induce the largest change in the classifier, aiming to estimate the influence of learning from a data point on the current model. Learning Active Learning (LAL, \cite{lal}) is a data-driven approach that predicts potential error reduction using classifier properties and data, but it may lack robustness to outliers.
Batch-mode Discriminative and Representative AL (BMDR, \cite{bmdr}) queries a batch of informative and representative examples by minimizing empirical risk bound, albeit restricted to binary cases. Query-By-Committee (QBC, \cite{qbc_class}) leverages a committee of classifiers to identify data points with the largest disagreement for labeling, maximizing the information gained from each labeled instance.

A recent survey \cite{survey_ijcai} bench-marking various active learning methods across diverse data sets, revealed that no single method uniformly outperforms others. It showed that the best active learning approach depends on the type of data set, and the classification problem.

In this paper, we propose a novel approach that integrates multiple steps to effectively leverage model-based active learning for the creation of high-quality training data sets. Our methodology begins with cold-start labeling based on random sampling. Additionally, we introduce a classification tree to detect regions where additional labels should be queried, updating previous active learning methodologies developed for regression tasks \cite{ashna}, to address classification challenges. Our method, namely Classification Tree-based Active Learning (CT-AL), exploits the information of labeled samples to partition the input-output space into regions that are homogeneous in the output. The new samples to be queried are selected from regions with high density of unlabeled samples, and from regions with high entropy. Additionally, diversity and representativity-based criteria are also used to to further enhance the sampling process.
Extensive experiments are conducted to illustrate our method and to compare it with state-of-the-art methods, on data sets of various sizes. Although many AL methods are restricted to certain kinds of data sets, either balanced or imbalanced, binary or multi-class, our work shows that CT-AL succeeds to outperform random sampling and other AL methods for all types of data sets, irrespective of its dimensions, imbalances and number of classes. The main contributions of our work are summarised as follows:

\begin{itemize}

\item We propose a model-based AL approach for classification based on classification trees.

\item Using diversity and representativity-based criteria, we show that the sampling from different regions of the leaves can be enhanced substantially.

\item We demonstrate, with experiments on various benchmark datasets, how a classification tree in the joint input-output space can detect informative samples to enhance the accuracy of the learned estimator by taking the output into account, especially on imbalanced data sets.

\end{itemize}

The remainder of this paper is structured as follows: Section \ref{sec:method} introduces our proposed methodology, detailing each step in the active learning process. In Section \ref{sec:experiments}, ablation studies and experimental results are presented to illustrate the performance of our approach compared to the state of the art. Finally, a discussion and conclusion is provided in Section \ref{sec:conc}, summarizing our findings and outlining directions for future research.

\section{Method}\label{sec:method}
\begin{figure}[t]
    \centering
    \includegraphics[width = \textwidth]{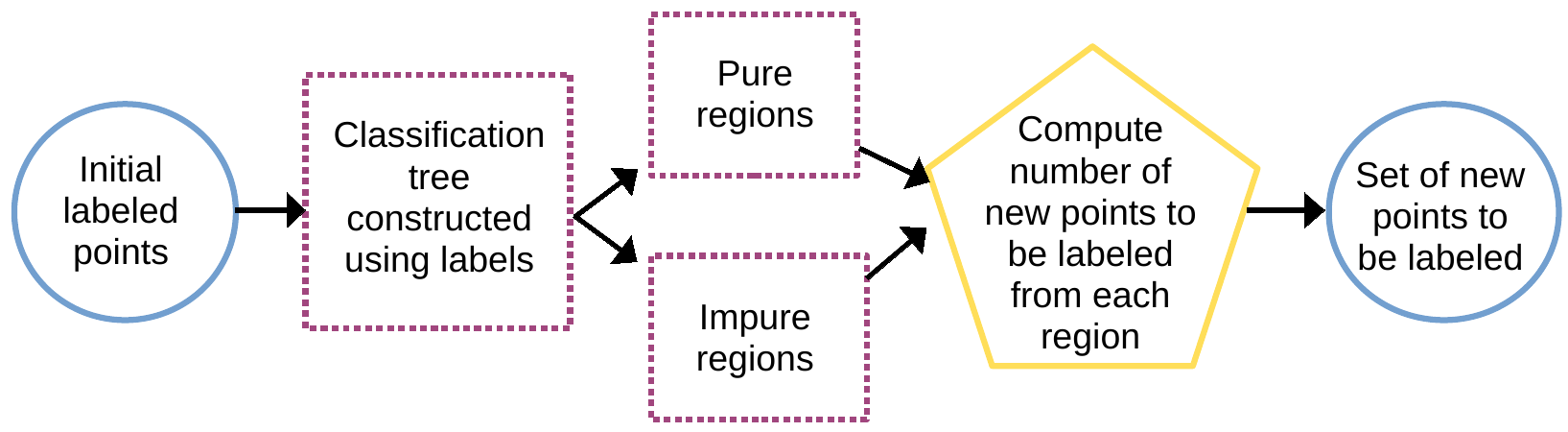}
    \caption{Flowchart of the proposed method. Blue circles correspond to model-free steps, while magenta dashed rectangles correspond to model-based steps. The yellow pentagon defines the criteria to query new samples, which is detailed in Section \ref{sec:ct-al}. The whole budget $n$ is divided into the first initial points, $n_{\text{init}}$, and the new points to be labeled, $n_{\text{act}}$.}
    \label{fig:flowchart}
\end{figure}

In this section, we formally introduce the context of AL in classification, and more specifically model-based AL. Then, we introduce our main method, classification tree-based active learning (CT-AL). This is decomposed into 2 parts: decomposition of the space into homogeneous regions, and sub-sampling from each region. 
 A flowchart describing the whole process is shown in Figure \ref{fig:flowchart}. 

\subsection{Active learning in multi-class classification} \label{sec:ct-al}
We consider a multi-class classification problem such that the input space is $\mathcal X \subset \mathbb{R}^D$ and the output space $\mathcal{Y}=\{1,\dots,c\}$ is a set of unknown classes of cardinal $c \in \mathbb{N}, c\geq 2$.
Let $\mathbf X\in \mathcal{X}$ be a $D$-dimensional random vector and let $Y \in \mathcal{Y}$ be a random variable linked to $\mathbf X$ through a classification model. 
In pool-based AL methods, a data set of size $N$ with all observations $\{{\mathbf{x}_{i}\}}_{i=1}^{N}$ unlabeled is given, and let $(y_i)_{i=1}^N$ denote the (unknown) associated labels. 

Formally, if $\hat{f}_{I_n}$ denotes the estimator among a class $\mathcal{F}$ of classification models learnt on a training set indexed by $I_n \subset \{1,\ldots,N\}$ of size $n$ with respect to the risk $R$,
\begin{equation} \label{eq:f_estimate}
 \hat{f}_{{I}_n} = \underset{f \in \mathcal{F} }{\operatorname{argmin}} \left\{\frac{1}{n} \sum_{i \in I_n} R(\hat{f}_{I_n}(\mathbf{x}_i), y_i) \right\},
\end{equation}

we look for the set of observations indexed by $\mathcal{I}_n \subseteq \{1,\ldots, N\}$ of size $n$ such that the transductive risk is minimized:

\begin{equation}\label{eq:In}
\mathcal I_n = \underset{I_n \subseteq \{1,\ldots,N\} }{\operatorname{argmin} } \left\{\frac{1}{N-n} \sum_{i \notin I_n} R(\hat{f}_{I_n}(\mathbf{x}_i), y_i)\right\}.
\end{equation}

However, in practice, one does not have access to the transductive risk because the labels are not observed. 

\subsection{Decomposition of the space using a Classification tree}

\paragraph{Classification tree}

The method we propose is based on standard classification trees. Classification trees partition the input space into a set of $K$ hyper-rectangles, referred to as regions and denoted $\mathcal{R}_k = \prod_{\ell = 1}^p [a_{k,\ell} , b_{k,\ell}]$ for $1\leq k \leq K$, and use  the majority vote to assign a class to each region $k$:
 $$f(\mathbf{x};\Theta) = \sum_{k=1}^K c_k \mathbf{1}_{\{\mathbf{x}\in \mathcal{R}_k\}},$$
 where the set of parameters $\Theta = ((\mathcal{R}_k, c_k)_{1\leq k \leq K})$ corresponds to the set of regions and the associated prediction.  The best regions are constructed recursively by finding the best feature and the best splitting point to divide a current region into two sub-regions that makes the prediction less variable. 
 As the splitting process is dyadic, it can be represented as a tree, where each node determines the features to split and its corresponding value, and the final partition is given by the leaves of the tree. 

\paragraph{Pure and impure leaves}

Let $I_{\text{init}}$ be the indices of the first samples detected by random sampling. We construct a classification tree with $K$ leaves using the corresponding labeled set $(\mathbf x_{i}, y_i)_{i \in I_{\text{init}}}$, and use it to predict every unlabeled sample: $(\hat Y^{I_{\text{init}}}_i)_{i \notin I_{\text{init}}}$. As such, each leaf now consists of a few labeled samples, and many unlabeled samples. The leaves of the tree can be classified into 2 kinds of regions:

\begin{enumerate}
    \item[(a)] pure regions: where all the true labels belong to the same class;
    \item[(b)] impure regions: where labels differ.
\end{enumerate}

\begin{figure}[t]
\centering
    \includegraphics[width=0.7\textwidth]{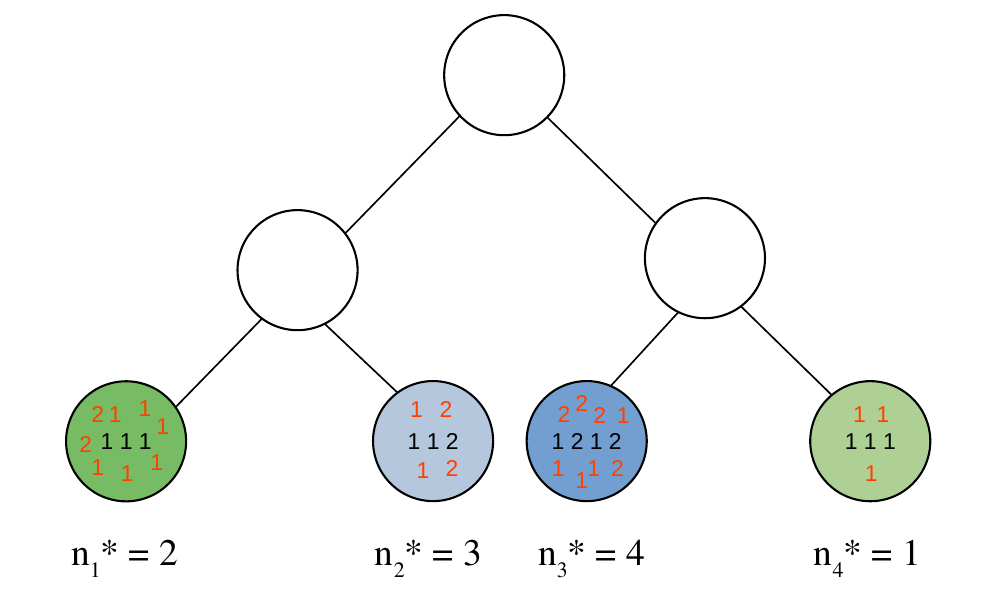}
    \caption{A classification tree learnt on labeled samples is depicted as an illustration of the different kind of nodes. Here, we consider the binary case, with classes 1 and 2. Leaves are classified as 'pure' when all the true labels in a regions belong to the same class, as depicted in green, and as 'impure' when there exist different classes among the labels, shown in blue. All true labels classes are shown as black, while unlabeled samples are shown as red numbers. The dark and light shades represent the high and low density regions of unlabeled samples, respectively. The $n_k^*$ below  each leaf are the number of new samples to be labeled from those regions using CT-AL.}
    \label{fig:CTAL_leaves}
\end{figure}

This is illustrated in Figure \ref{fig:CTAL_leaves}, where the green regions correspond to pure leaves while the blue are impure. Note that these regions are defined by the true labels only.
Since we trust the pure regions more than the impure ones, more labels must be sampled from impure regions, to be able to decompose the corresponding misunderstood region into several smaller pure regions. The budget, ${n_{act}}$, is then distributed into the different regions according to their nature. 

\paragraph{Cardinal of labels to be get in each leaf}
In regression, as demonstrated in \cite{mondrianTrees} and used in \cite{ashna}, the best number of points to be sampled in each region is proportional to the density of the region (number of unlabeled points) and the variance of the outputs in the leaf. 
To adapt this criterion to classification, entropy is used instead of the variance, and to identify regions that are pure and impure: impure regions will have non-zero entropy. The entropy, $S_k$ for each leaf $k$ is defined using Shannon information gain:

\begin{align} \label{eq:entropy}
S_k &= -\sum{p_{i,k} log(p_{i,k})}
\end{align}

where $p_{i,k}$ are the predicted probabilities of each class $i$ in leaf $k$.

Second, the density, corresponding to the number of unlabeled points, can be computed as follows:

\begin{align} \label{eq:pi}
\pi_k &= \frac {\vert i \notin I_{\text{init}}: \mathbf{x}_i \in \mathcal{R}_k \vert}{N}.
\end{align}

Conditionally to the first labeled set, we pick $n_k^*$ samples for labeling from leaf $k$, depending on the impurity of the leaf. First, the number of pure and impure leaves are counted. Since we want to sample more from impure leaves, this is ensured by distributing the budget into the two types of leaves as follows:

\begin{equation} \label{eq:sample_size_pure}
n_{pure} = \frac{n_{act}}{(1 + 3*\frac{max(1,n_{\text{impure leaves}})}{max(1,n_{\text{pure leaves}})})}
\end{equation}

where $n_{pure}$ are the number of samples to be labeled from the pure leaves, $n_{\text{impure leaves}}$ are the total number of impure leaves and $n_{\text{pure leaves}}$ are the total number of pure leaves. Consequently, the number of samples to be labeled from the impure leaves are $n_{act}$ - $n_{pure}$. Note, that this is one of the different possible ways to sample more from impure leaves, and is used as a proof of concept in this work.

Finally, the number of samples to be labeled from each leaf, $n_k^*$, is computed as:

\begin{equation} \label{eq:number_of_points}
n_k^* = n_{\text{act}} \frac{\sqrt{\pi_k E_k}}{\sum_{\ell=1}^K \sqrt{\pi_\ell E_\ell}};
\end{equation}

where $\pi_k$ denotes the probability that an unlabeled sample $\mathbf{x}_i$ belongs to leaf $k$, given in Equation \ref{eq:pi} and $E_k$ is given by:

\[
    E_k= 
\begin{cases}
    1  & \text{if } S_k =  0,\\
    S_k              & \text{otherwise},
\end{cases}
\]

where $S_k$ is the entropy of leaf $k$. Note that using $E_k$ = 1 does not imply high entropy for pure regions. Since $n_{act}$ has been distributed among pure and impure regions using equation \ref{eq:sample_size_pure}, the criteria to determine $n_k^*$ from the two types of regions are independent: using only the density of unlabeled samples to query new points from pure regions (where entropy is zero), while using both density of unlabeled samples and entropy among the pure labels to query new points from impure regions. An example of using CT-AL is illustrated in detail in Figure \ref{fig:CTAL_leaves}.

Using $n_k^*$, the $n_{\text{act}}$ samples to be labeled are thus divided into the sets $I_{\text{act}}^k$ for each leaf $k$, and the final labeled set proposed to approximate $\mathcal{I}_n$ from Eq. \eqref{eq:In} is 
$$\hat{\mathcal{I}}_n = I_{\text{init}} \cup (\cup_{k=1}^K I_{\text{act}}^k).$$

The $n_{\text{act}} = n - n_{\text{init}}$ samples can be detected sequentially (retraining the model after adding each sample), they can be split into batches of moderate dimension or be detected in one step.
The method is summarized in the form of a pseudo-code in Algorithm \ref{alg:two}. 
This AL design works well as it is but can be improved further by splitting the active part into various steps, i.e.~by adding few samples at each step, followed by retraining the classification tree at each iteration. This process is then repeated till the desired size of the training set is reached.

\begin{algorithm}[t]

\caption{Classification Tree-based Active Learning: CT-AL (RS)}\label{alg:two}

\textbf{Input}: Labeled set $(\mathbf x_i, y_i)_{i \in I_{\text{init}}}$ and unlabeled set $(\mathbf x_i)_{i\notin I_{\text{init}}}$;\\
$n_{\text{act}}$ the maximum number of new samples to be labeled

\begin{algorithmic}[1]

 \STATE Construct a standard classification tree with $K$ leaves using $(\mathbf x_i, y_i)_{i \in I_{\text{init}}}$

\STATE{Compute $n_{pure}$ and $n_{impure}$ using Eq. \eqref{eq:sample_size_pure}}

\FOR{$k = 1,\ldots,K$}

\STATE{Compute $S_{k}$ and $\pi_{k}$ using Eqs. \eqref{eq:entropy} and \eqref{eq:pi}}

\STATE{Calculate the number of samples $n_k^*$ to be labeled from leaf $k$ using Eq. \eqref{eq:number_of_points}}

\STATE{Detect $I_{\text{act}}^k$, the set of $n_k^*$ observations from leaf }

\ENDFOR
\end{algorithmic}

 \textbf{Output}: The set $\cup_{k=1}^K (\mathbf x_i)_{i \in I_{\text{act}}^k}$ of observations to be labeled, labeled from each leaf randomly
\end{algorithm}

\subsection{Diverse and representative sampling in each region}\label{subsec:div-rep}

Once $n_k^*$ is computed for each leaf, the $I_{\text{act}}^k$ samples need to be queried from each region. These new points can be queried randomly as suggested by \cite{mondrianTrees}. This would increase the randomness of the selected samples. This can be improved by taking advantage of the input-space information that is known for all samples, and defining a criterion based on the diversity and representativeness the inputs. Although such criteria have been studied in regression tasks \cite{ashna}, their application in classification has not been investigated. We describe the criteria in detail below.

We denote $J_1 = \{1, \ldots, n_1^*\}$ and $J_{k} = \{\sum_{u = 1}^{k-1} n^*_{u}+1, \ldots, \sum_{u = 1}^{k} n^*_{u}\}$ for $k \in \{2,\ldots, K\}$. 
First, a k-means clustering is performed in each leaf $k$ on the unlabeled samples, with $n_k^*$ clusters, denoted by $(c_\ell)_{\ell\in J_k}$. The sample closest to the centroid of these clusters, denoted $(\mathbf{x}_{j_{\ell}^*} )_{\ell \in J_k}$, is a good representative of these clusters.
However, all the $n_k^*$ centroids may not form a diverse enough set of samples in leaf $k$, and more generally, the set of all $n_{\text{act}} = \sum_{k=1}^K n_k^*$ centroids may not form a diverse enough set of samples in the feature space, amongst themselves and also with respect to the samples that have initially been labeled. Therefore, for each cluster $\ell = 1,\ldots, n_{\text{act}}$, we update the centroids by optimising the following criteria,
\begin{equation}
\label{bt(irdm3)}
j^*_\ell = \underset{j \in c_\ell}{\operatorname{argmax}} [\Delta(\mathbf x_j) - R(\mathbf x_j)].
\end{equation}

The criteria above considers the diversity in the term $\Delta(\mathbf x_j)$ and ensures representativeness in the term $R(\mathbf x_j)$, where for each unlabeled sample $\mathbf x_j \in c_{\ell}$, for $\ell= 1,\ldots, n_{\text{act}}$,

\begin{algorithm}[t]
\caption{Diverse and representative query in leaves: CT-AL(div-rep)}\label{alg:bt(irdm)}

\textbf{Input}: Labeled set $(\mathbf x_i, y_i)_{i \in I_{\text{init}}}$ and unlabeled set $(\mathbf x_i)_{i\notin I_{\text{init}}}$; $n_{\text{act}}$ the maximum number of new samples to be labeled

\begin{algorithmic}[1]

\item Construct a classification tree with $K$ leaves using $(\mathbf x_i, y_i)_{i \in I_{\text{init}}}$

\STATE{Compute $n_{pure}$ and $n_{impure}$ using Eq. \eqref{eq:sample_size_pure}}

\FOR{$k = 1,\ldots, K$}

\STATE{Compute $S_{k}$ and $\pi_{k}$ using Eqs. \eqref{eq:entropy} and \eqref{eq:pi}}

\STATE{Calculate the number of samples $n_k^*$ to be labeled from leaf $k$ using Eq. \eqref{eq:number_of_points}}

\STATE{Perform a k-means clustering in leaf $k$ with $n_k^*$ clusters} 

\STATE{Calculate $R$ using Eq. \eqref{bt(irdm1)}}
\ENDFOR

 \WHILE{it has not converged}

\FOR{$\ell = 1,\ldots,n_{\text{act}}$}
 
\STATE{Update $\Delta$ using the current centroids $(\mathbf{x}_{j_l^*})_{l \neq \ell}$}

\STATE{Use Eq. \eqref{bt(irdm3)} to find the next sample to be labeled $\mathbf{x}_{j_\ell^*}$}

\ENDFOR
\ENDWHILE
\item \textbf{Output}: the set {$\cup_{\ell=1}^{n_\text{act}} \mathbf x_{j^*_\ell}$ of} observations to be labeled

\end{algorithmic}
\end{algorithm}

\begin{align}
\label{bt(irdm1)}
R(\mathbf x_j) &= \frac{1}{\vert c_\ell\vert - 1} \sum_{\mathbf x_m \in c_\ell \atop m \neq j}\|\mathbf x_{j} - \mathbf x_{m}\|;\\
\label{bt(irdm2)}
\Delta(\mathbf x_j) &= \min_{m \in I_{\text{init}}\cup \{j_l^* \}_{l \neq \ell}} {\|\mathbf x_{j} - \mathbf x_{m}\|} = \min_{m \in I_{\text{init}}\cup \{j_l^*\}_{l \neq \ell}} {d_{jm}}.
\end{align}

$\Delta(\mathbf x_j)$ includes information of the cluster centroids of the clusters in leaf $k$, and also of the cluster centroids in all other leaves, and all the samples that were labeled beforehand. Thus, this set of samples is much more informative, than randomly sampling from the leaves.

The above routine is then repeated for all other clusters in all leaves and is repeated until the $n_{act}$ samples to be selected are optimised (or till a threshold is reached). Thus, we select a set of samples that are diverse and representative of the feature space. Note that through the classification tree, we also take the outputs into account. We show in Section \ref{sec:experiments} how wrapping CT-AL with this criterion works efficiently in various settings. The pseudo-code of the algorithm is provided in Algorithm \ref{alg:bt(irdm)}.

\section{Experiments}\label{sec:experiments}
In this section, we first introduce the data sets and the experimental setup. Then, we conduct an ablation study to understand the benefit using the classification tree in active learning, as well as the advantage of the wrapper. Finally, we compare the performance of our method with the state-of-the-art. 

\subsection{Data sets}
We carry out experiments on data sets that are frequently used in active
learning. Table \ref{tab:datasets} presents statistics of these data sets, showing the data set size, dimension of the inputs, number of classes and class imbalance ratio. The Diabetes, Statlog (German) and Nursery data sets are available in the UCI machine learning repository (\href{https://archive.ics.uci.edu/}{archive.ics.uci.edu}). The references for other data sets are provided in the table. These data sets were chosen to accommodate a wide variety of properties: data sets that are very balanced, while others that are highly imbalanced, data sets with a wide range of number of samples and features, and data sets with different number of classes.

\begin{table}[t]
\setlength{\tabcolsep}{8pt} 
\renewcommand{\arraystretch}{1.2} 
\caption{Benchmarking data sets used in this work. D, N, K are the feature dimension, number of samples, and number of classes, respectively. The Imbalance Ratio (IR) is the ratio of the number of samples in the majority class to that of the minority class.}\label{tab:datasets}
\centering
\begin{tabular}{ccccc} \hline

Data set     & IR    & D & N & K  \\ \hline
Diabetes         & 1.87  & 8 & 768 & 2            \\
Statlog (German)     & 2.33 &  20 & 1000 & 2             \\
Banknote \cite{banknote}   &   0.83 &  4 & 1372 & 2           \\
Coil-20 \cite{coil20} & 1.00  & 1024 & 1440 & 20          \\
Phoneme & 2.41 &  5 & 5405 & 2     \\ 
Nursery \cite{banknote} & 0.09 &  8 & 12960 & 4     \\  \hline
\end{tabular}
\end{table}

\subsection{Experimental Setup}

For each data set, 20 samples were labeled initially using random sampling. This is followed by construction of the classification tree. The hyperparameters were chosen as follows: the minimum samples in the leaf of the decision tree was set to 10 (i.e. the trees do not branch if the number of samples in a leaf is less than 10), for meaningful entropy calculations, while simultaneously optimising the tree as much as possible. The quality of the splitting was controlled using the entropy criteria. We consider a test set of size 20\% of the whole data set, uniformly drawn, on which the final classifier inferred on the constructed training set is tested. The remaining 80\% was considered as the training pool from which the new samples were queried.

We compare our method to classical random sampling, that has been shown to be very competitive in active learning scenarios in classification tasks. We also compare CT-AL to two AL methods: the model-free method iRDM \cite{irdm}, and the model-based method QBC.
iRDM uses $k$-means clustering to partition the feature space into a number of clusters corresponding to the number of samples to be labeled. The samples closest to the centroids of these clusters are chosen as the starting points, which are optimised using feature space diversity. The samples selected are thus representative and diverse. QBC selects samples with the highest variance/entropy among the predictions from a committee of models. The committee is constructed by bootstrapping on the initial set of randomly labeled samples. The Scikit-active ML \cite{skactiveml2021} package was used to implement QBC, with decision trees in the committee of models. Codes for our method, and implementations of other methods were done in python, and will be provided in a public repository in the future.
Note that our method, RS and QBC are sequential i.e. new samples are added to the training set iteratively, while iRDM is not sequential. It gives exact number of sample for a given budget, and discards the previous samples for a new budget.

Balanced classification accuracy \cite{bal_acc} of the output was used as the performance measure for all the methods. It is defined as the average of the true positive rate and the true negative rate, i.e.

\begin{align}
\text{Balanced Accuracy} = \frac{((TP/(TP+FN)+(TN/(TN+FP))}{2}
\end{align}

where TP, TN, FP and FP are the number of true positives and negatives, and false positives and negatives, respectively.

The final classifier that we use to evaluate the method at each step is a random forest classifier built on the labeled set selected by the method. The number of trees in the forest were set to 50, and the splitting was controlled using entropy. The depth of the trees in the forest was defined by minimum samples in the leaf, set to 3, so as to avoid over-fitting. Note, as we compare different methods throughout the prediction phase, we keep the same hyperparameters for all the methods for consistency. Random forest was chosen as the final predictor because it is versatile and robust to outliers. It is especially appropriate for our approach because we build our training set using its fundamental architecture, which is classification trees. 
The experiments were run 100 times each for different train-test splits to compute statistics, and we discuss the relevance of differences based on a Wilcoxon rank sum test with p < 0.05 \cite{stattest}.
To see the evolution of the performance of model-based AL methods with respect to the number of labeled points, we do the following:
random sampling is used to choose the first 20 samples that need to be labeled. To maximize the benefit of the model-based portion of the technique, this quantity is kept low. Following the labeling of each of the 20 samples, the active step is carried out. The model is then fitted when the additional samples are added to the training set. 

\subsection{Ablation study}

\begin{figure}[H]
\centering
\includegraphics[scale=0.23]{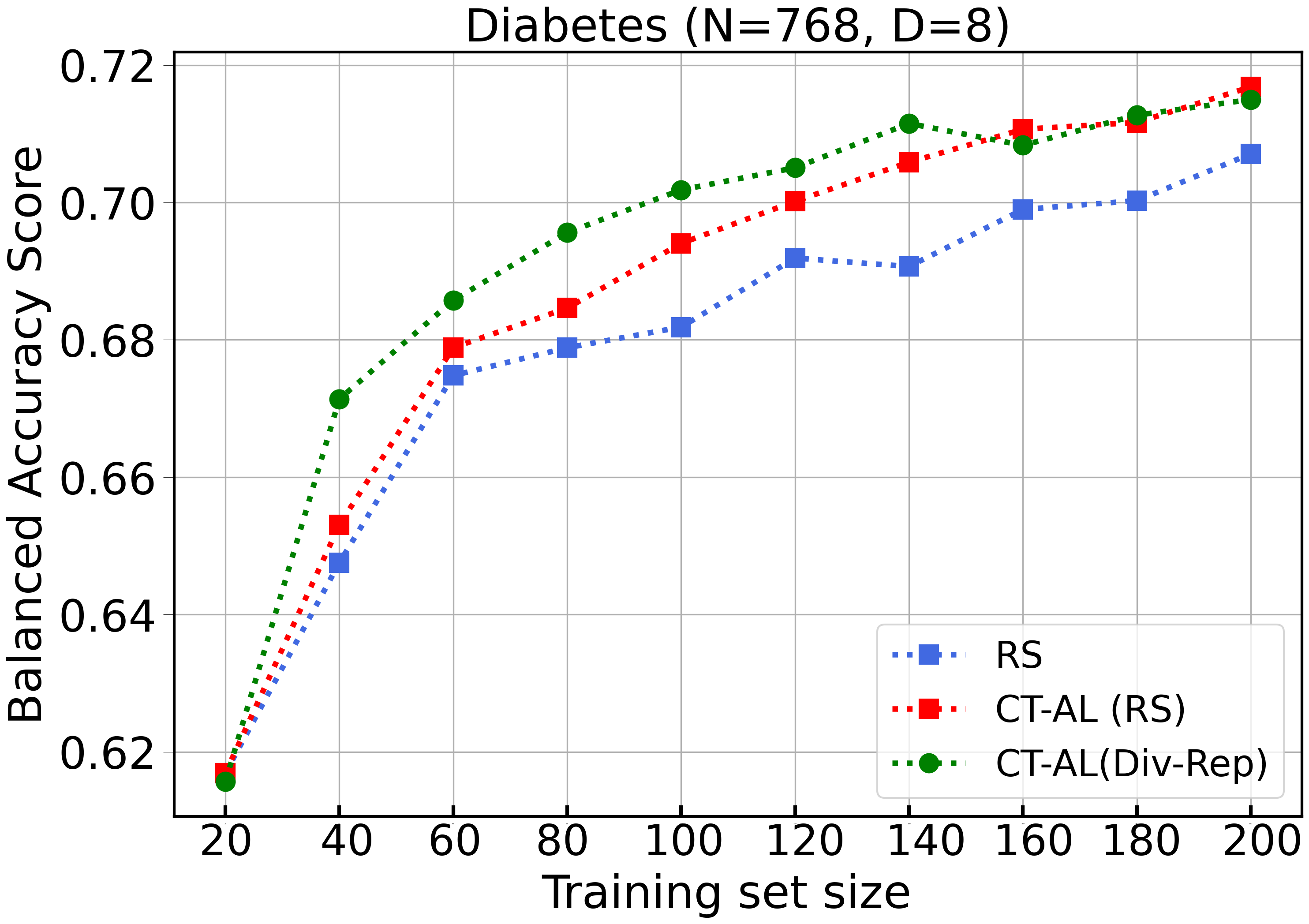} \hspace{0.1cm}
\includegraphics[scale=0.23]{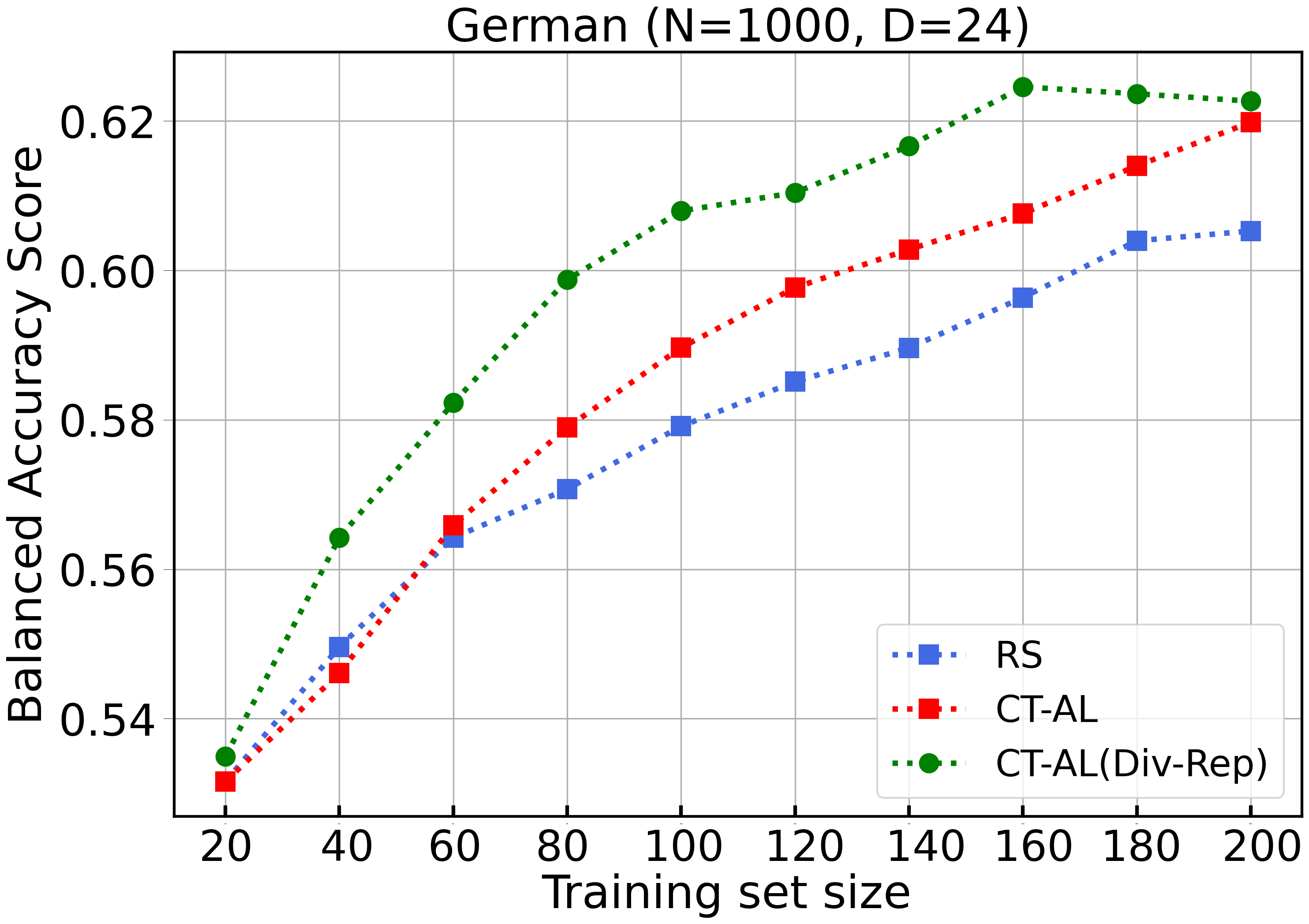}\\ \vspace{0.25cm}
\includegraphics[scale=0.23]{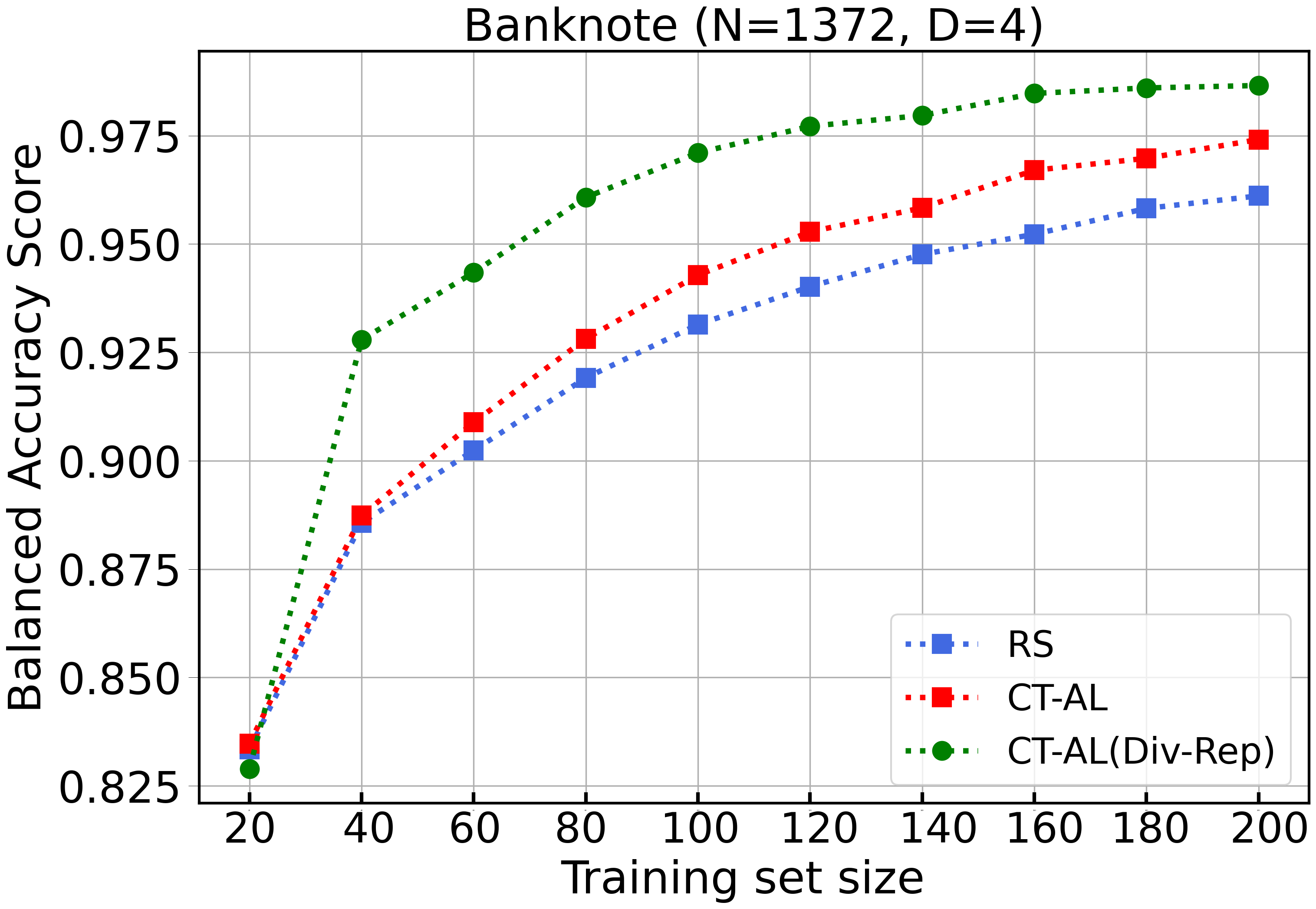} \hspace{0.17cm}
\includegraphics[scale=0.23]{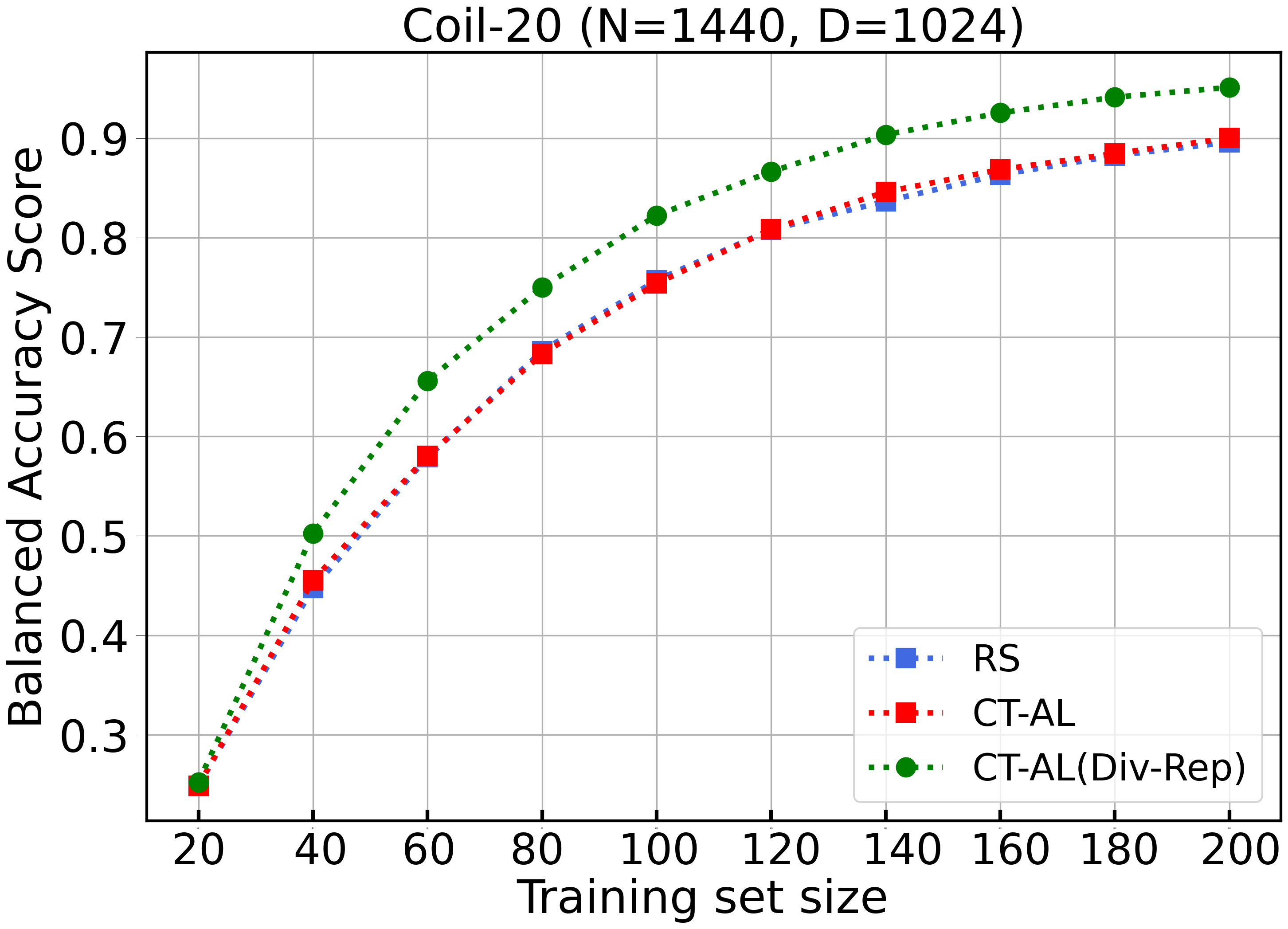}\\ \vspace{0.25cm}
\includegraphics[scale=0.23]{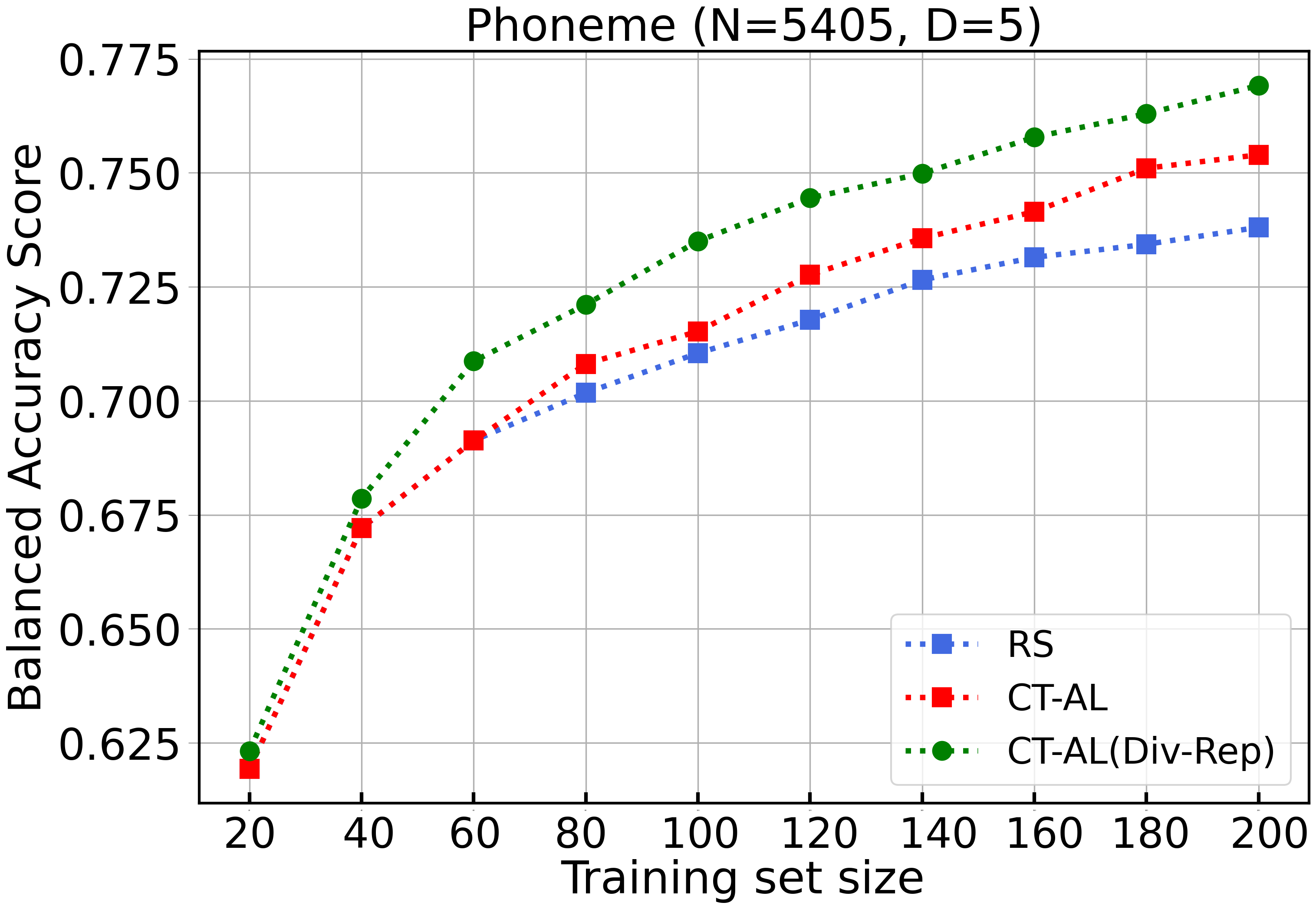} \hspace{0.1cm}
\includegraphics[scale=0.23]{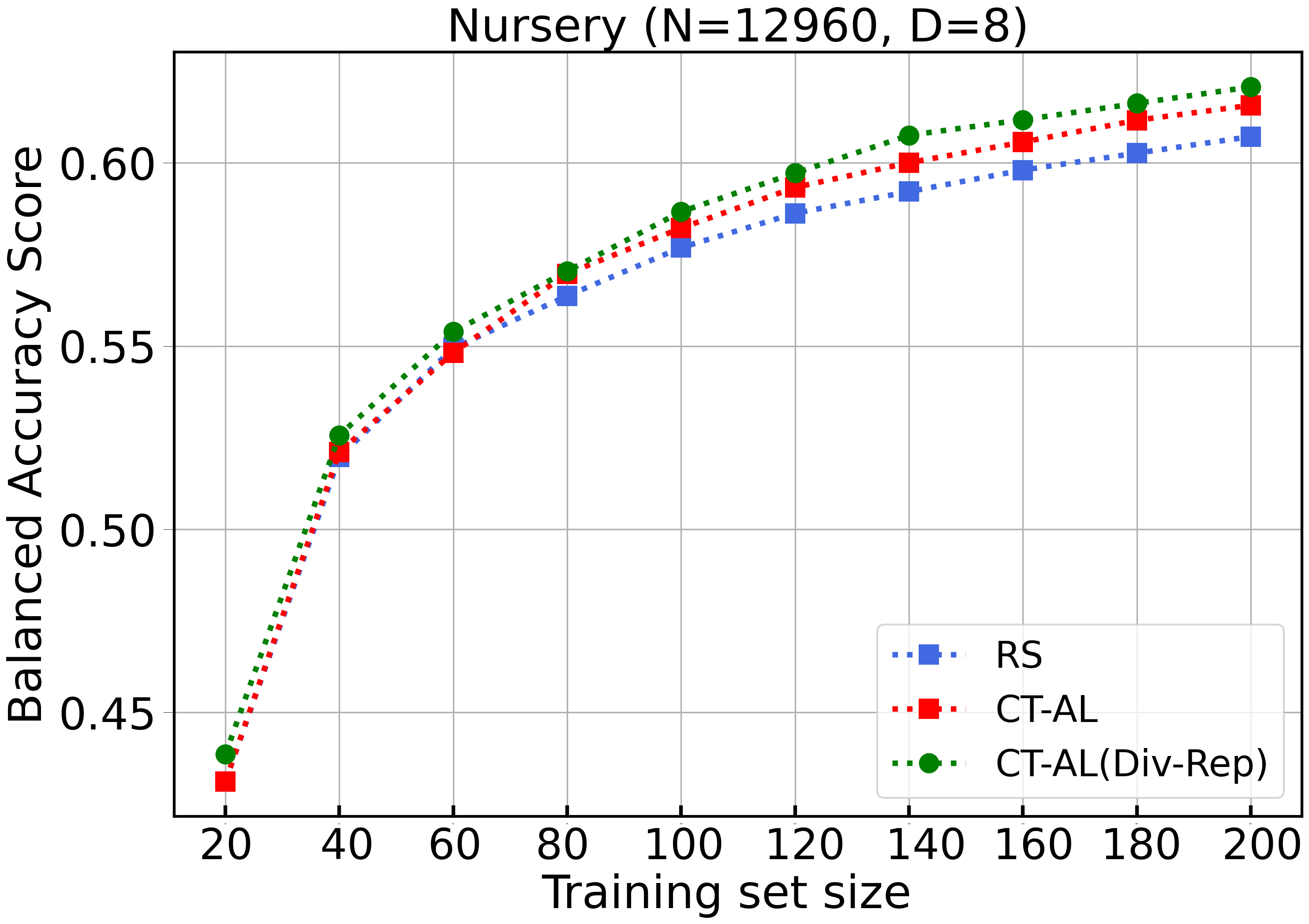}

\caption{Performance in prediction using balanced accuracy score, averaged over 100 runs for different train-test splits, when the training set is constructed using random sampling (RS) and CT-AL using different criteria to sample from the leaf regions, for 6 different data sets. CT-AL with random sampling from the leaves is shown as CT-AL (RS), while CT-AL with diversity-representativity criterion is shown as CT-AL (div-rep). The training set size varies from 20 to 200.}
\label{fig:ablation}
\end{figure}

Ablation studies were performed on the 6 data sets described in the previous section. In Figure \ref{fig:ablation} we show the evolution of balanced classification accuracy by using samples selected by RS, CT-AL and CT-AL(div-rep). For CT-AL with random sampling in leaves and CT-AL(div-rep), 20 samples are added in each iteration, and the classification tree used to select the next set of samples is retrained. Random forest classifiers are trained to obtained these curves, as mentioned before.

The results show a distinct difference in balanced accuracy among the three methods. 
We first investigate the case where samples are chosen randomly from the leaves. In general, using the regions defined by the classification tree leads to more informative samples than RS as can be seen from the red curve outperforming the blue. For Coil-20 data set, RS and CT-AL with RS in the leaves give similar accuracy. This is due to the fact that this data set is the only completely balanced data set among the 6. Since the tree-based structure focuses on the outputs more than the inputs, using the tree for active learning, while randomly sampling from the leaves has minimal benefit in terms of accuracy when the data set is perfectly balanced. On the other hand, for all other data sets, class imbalance is high. Thus, our criteria to sample more from diverse regions of the outputs ensures samples from all classes. This is why using the partitions of the tree to query new points is beneficial, and leads to a significantly higher accuracy. This technique of sampling is very important in the case of imbalanced data sets, where learning accurate models with low data is known to be an extensively challenging task. 

Further, on choosing the samples from the leaves using the diversity - representativity based criteria defined in Section \ref{subsec:div-rep}, we note that the accuracy improves further. This is a result of the fact that taking the input-space structure of the data into account is necessary as well, and diversifying the samples chosen from the leaves leads to an even more accurate model. Note that in all cases, CT-AL(div-rep) improves random sampling significantly. As the data sets are of very different nature, this shows that CT-AL(div-rep) is robust and can be reliable AL method for real-world problems. Thus, in general, CT-AL (div-rep) proves to be an apt method for AL in classification.

\subsection{Experimental Results}

We now compare our method to some common and competitive AL methods, discussed previously. The performance of each method for 100 and 200 labeled samples is depicted in Table \ref{tab:perf} using balanced accuracy score (average and standard deviation (shown in parenthesis) over 100 repetitions). The best performer is highlighted with bold and star.

First, we see that RS and QBC perform very poorly, giving low values of accuracy for all the data sets. QBC also has the highest standard deviation among all the methods, which is not ideal for active learning. Moreover, for banknote, coil-20 and nursery data sets, QBC performs worse than RS. This shows that classical random sampling, even though does not use any information of the unlabeled samples, tends to sample better than AL methods in many cases. 

Analysing the performance of CT-AL, for both the training set sizes shown, it can be seen that CT-AL with the diversity and representativity maximisation criteria in the leaves is the best performer in almost all the cases. In data sets of smaller sizes, iRDM and QBC are competitive (shown by statistical equivalent result in the table, according to the Wilcoxon rank sum test, with p < 0.05 \cite{stattest}), although CT-AL still gives the highest accuracy. This highlights the significance of clustering-based sampling in classification. iRDM also has the advantage of starting from scratch at every budget, implying it selects samples given a fixed budget, while CT-AL samples sequentially, and does not discard previously selected points, still giving a much better performance. As in a real situation, one may not have a fixed budget but a certain threshold of accuracy as the stopping criterion, using CT-AL would be more apt.

\begin{table}[H]
\centering
\caption{Performance of each method for 100 and 200 labeled samples, depicted in using balanced accuracy score (average and standard deviation (shown in parenthesis) over 100 repetitions). Best performer is highlighted with bold and star. Its statistical equivalent result, according to the Wilcoxon rank sum test with p < 0.05 \cite{stattest}, is also shown in bold.}\label{tab:perf}
\setlength{\tabcolsep}{6.8pt} 
\renewcommand{\arraystretch}{1.25} 
\begin{tabular}{c|c|c|c|c|c|c}
\hline   
                                                                            & \begin{tabular}[c]{@{}c@{}}Training \\ set size\end{tabular} & RS                           & iRDM                         & QBC                          & \begin{tabular}[c]{@{}c@{}}CT-AL \\ (RS)\end{tabular} & \begin{tabular}[c]{@{}c@{}}CT-AL\\  (div-rep)\end{tabular} \\ \hline
\multirow{4}{*}{Diabetes}                                                   & \multirow{2}{*}{100}                                         & 0.682                        & \textbf{0.701}               & \textbf{0.698}               & \textbf{0.694}                                        & \textbf{0.702*}                                            \\
                                                                            &                                                              & \multicolumn{1}{l|}{(0.046)} & \multicolumn{1}{l|}{(0.044)} & \multicolumn{1}{l|}{(0.035)} & \multicolumn{1}{l|}{(0.040)}                          & \multicolumn{1}{l}{(0.039)}                               \\
                                                                            & \multirow{2}{*}{200}                                         & 0.707                        & \textbf{0.716}               & \textbf{0.712}               & \textbf{0.717*}                                       & \textbf{0.715}                                             \\
                                                                            &                                                              & \multicolumn{1}{l|}{(0.032)} & \multicolumn{1}{l|}{(0.034)} & \multicolumn{1}{l|}{(0.029)} & \multicolumn{1}{l|}{(0.032)}                          & \multicolumn{1}{l}{(0.035)}                           \\ \hline   
\multirow{4}{*}{\begin{tabular}[c]{@{}c@{}}Statlog\\ (German)\end{tabular}} & \multirow{2}{*}{100}                                         & 0.579                        & 0.588                        & \textbf{0.596}               & 0.590                                                 & \textbf{0.608}                                             \\
                                                                            &                                                              & \multicolumn{1}{l|}{(0.046)} & \multicolumn{1}{l|}{(0.035)} & \multicolumn{1}{l|}{(0.047)} & \multicolumn{1}{l|}{(0.041)}                          & \multicolumn{1}{l}{(0.043)}                               \\
                                                                            & \multirow{2}{*}{200}                                         & 0.605                        & 0.613                        & \textbf{0.619}               & \textbf{0.620}                                        & \textbf{0.623*}                                            \\
                                                                            &                                                              & \multicolumn{1}{l|}{(0.035)} & \multicolumn{1}{l|}{(0.038)} & \multicolumn{1}{l|}{(0.042)} & \multicolumn{1}{l|}{(0.034)}                          & \multicolumn{1}{l}{(0.040)}                            \\ \hline  
\multirow{4}{*}{Banknote}                                                   & \multirow{2}{*}{100}                                         & 0.931                        & \textbf{0.967}               & 0.921                        & 0.943                                                 & \textbf{0.971*}                                            \\
                                                                            &                                                              & \multicolumn{1}{l|}{(0.028)} & \multicolumn{1}{l|}{(0.015)} & \multicolumn{1}{l|}{(0.038)} & \multicolumn{1}{l|}{(0.027)}                          & \multicolumn{1}{l}{(0.016)}                               \\
                                                                            & \multirow{2}{*}{200}                                         & 0.961                        & 0.981                        & 0.938                        & 0.974                                                 & \textbf{0.987*}                                            \\
                                                                            &                                                              & \multicolumn{1}{l|}{(0.019)} & \multicolumn{1}{l|}{(0.010)} & \multicolumn{1}{l|}{(0.036)} & \multicolumn{1}{l|}{(0.014)}                          & \multicolumn{1}{l}{(0.008)}                               \\  \hline   
\multirow{4}{*}{Coil-20}                                                    & \multirow{2}{*}{100}                                         & 0.757                        & 0.716                        & 0.700                        & 0.754                                                 & \textbf{0.822*}                                            \\
                                                                            &                                                              & \multicolumn{1}{l|}{(0.043)} & \multicolumn{1}{l|}{(0.044)} & \multicolumn{1}{l|}{(0.053)} & \multicolumn{1}{l|}{(0.047)}                          & \multicolumn{1}{l}{(0.041)}                               \\
                                                                            & \multirow{2}{*}{200}                                         & 0.896                        & 0.907                        & 0.883                        & 0.900                                                 & \textbf{0.951*}                                            \\
                                                                            &                                                              & \multicolumn{1}{l|}{(0.28)}  & \multicolumn{1}{l|}{(0.042)} & \multicolumn{1}{l|}{(0.029)} & \multicolumn{1}{l|}{(0.030)}                          & \multicolumn{1}{l}{(0.020)}                               \\   \hline   
\multirow{4}{*}{Phoneme}                                                    & \multirow{2}{*}{100}                                         & 0.710                        & 0.709                        & 0.708                        & 0.715                                                 & \textbf{0.735*}                                            \\
                                                                            &                                                              & \multicolumn{1}{l|}{(0.042)} & \multicolumn{1}{l|}{(0.035)} & \multicolumn{1}{l|}{(0.042)} & \multicolumn{1}{l|}{(0.033)}                          & \multicolumn{1}{l}{(0.037)}                               \\
                                                                            & \multirow{2}{*}{200}                                         & 0.738                        & 0.754                        & 0.746                        & 0.754                                                 & \textbf{0.770*}                                            \\
                                                                            &                                                              & \multicolumn{1}{l|}{(0.029)} & \multicolumn{1}{l|}{(0.026)} & \multicolumn{1}{l|}{(0.031)} & \multicolumn{1}{l|}{(0.023)}                          & \multicolumn{1}{l}{(0.026)}                               \\  \hline   
\multirow{4}{*}{Nursery}                                                    & \multirow{2}{*}{100}                                         & \textbf{0.577}               & 0.570                        & 0.513                        & \textbf{0.582}                                        & \textbf{0.584*}                                            \\
                                                                            &                                                              & \multicolumn{1}{l|}{(0.066)} & \multicolumn{1}{l|}{(0.068)} & \multicolumn{1}{l|}{(0.082)} & \multicolumn{1}{l|}{(0.068)}                          & \multicolumn{1}{l}{(0.064)}                               \\
                                                                            & \multirow{2}{*}{200}                                         & 0.607                        & 0.604                        & 0.552                        & \textbf{0.616}                                        & \textbf{0.618*}                                            \\
                                                                            &                                                              & \multicolumn{1}{l|}{(0.068)} & \multicolumn{1}{l|}{(0.063)} & \multicolumn{1}{l|}{(0.084)} & \multicolumn{1}{l|}{(0.071)}                          & \multicolumn{1}{l}{(0.067)}                              \\  \hline   
\end{tabular}
\end{table}

The results also show that CT-AL(div-rep) has a much larger accuracy compared to other approaches for (i) multi-class data sets and (i) imbalanced data sets. This is in line with our understanding of CT-AL which ensures sampling from all regions of the output space, thereby reducing the class imbalance in the samples chosen from the data set. As most real world data sets are highly imbalanced in nature, the use of an active learning method like CT-AL for such data sets is immensely relevant, and is one of the main contributions of our work. We would also like to highlight that CT-AL with random sampling is leaves is also competitive to CT-AL(div-rep) as shown in Table \ref{tab:perf}. This implies that the regions defined by the classification, and the subsequent sampling from pure and impure regions are of importance. Even though CT-AL(RS) using RS to sample from the leaves, it outperforms RS, QBC and iRDM in most of the cases.

Table \ref{tab:perf} also shows the standard deviation for all the methods over the 100 train-test splits. As can be seen, the standard deviation of CT-AL is lowest, implying that it is the least affected by the training pool of samples. This makes CT-AL a reliable and high confidence method since for low data settings in a real case, one does not apply the method many times, rather only once. The training curves for all the methods for training set sizes from 20 to 200 are provided in the appendix.

\section{Discussion and Conclusion} \label{sec:conc}

In this work, we propose a model-based active learning method for multi-class classification that uses the knowledge available in the data to query diverse, representative and informative samples.  By selecting new samples from each leaf of the tree, which is constructed on the labeled part of the data set, the new samples contain information of both the input and the output. Using the knowledge of regions with high entropy and high density, the budget to be labeled is distributed in those regions such that more queries are labeled from the impure regions. 

To further improve the algorithm, diversity and representativity-based criteria are defined and used to sample form the regions of the tree. Tests on several benchmark data sets show that our approach works significantly better than random sampling, and other competitive AL methods.
We also highlight the low variance of the results on all the experiments made in this study, which is vital when carrying out active learning.

Since the method is model-based, further improvements are possible regarding how the model, that selects new points, can be trained. In the future, we would like to take advantage of semi-supervised learning in active learning scenarios, for instance by training the classification tree with true as well as pseudo labels, thus training on a much larger sample size. This could potentially improve the model accuracy further, and be beneficial for the low data regimes.

In perspective, integrating active learning with transfer learning techniques could extend the applicability of our approach to scenarios where labeled data is scarce in the target domain but abundant in related source domains. Leveraging knowledge transfer from a source domain could aid in selecting informative samples for labeling in the target domain, thereby accelerating the learning process. Furthermore, exploring ensemble methods within the active learning framework could offer benefits in terms of robustness and diversity of sampled points. Ensemble techniques such as bagging or boosting could be employed to aggregate multiple's model decisions when selecting samples, leading to more reliable and diverse queries. 

Finally, extending the evaluation of our approach to more challenging domains, such as noisy environments, would provide insights into its robustness and generalization capabilities.

\begin{credits}
\subsubsection{\ackname} 

We acknowledge the CINES, IDRIS and TGCC under project No. INP2227/72914/gen7211, as well as CIMENT/GRICAD for computational resources. This work has been partially supported by MIAI@Grenoble Alpes (ANR-19-P3IA-0003). The authors thank Valérie Monbet and Lies Hadjadj for their insights and discussions. Discussions within the active learning community at ECML PKDD 2023 are also acknowledged.

\subsubsection{\discintname}
The authors have no competing interests to declare. 
\end{credits}

\bibliographystyle{splncs04}
\bibliography{references_active}

\end{document}